\documentclass{article}

\usepackage{amsfonts}
\usepackage{amsmath}
\usepackage{amssymb}
\usepackage{amsthm}
\usepackage{graphicx}
\usepackage{booktabs}
\usepackage[margin=0.9in]{geometry}
\usepackage{authblk}

\DeclareRobustCommand\vec[1]{\mathchoice{\mbox{\boldmath$\displaystyle#1$}}
{\mbox{\boldmath$\textstyle#1$}}
{\mbox{\boldmath$\scriptstyle#1$}}
{\mbox{\boldmath$\scriptscriptstyle#1$}}}

\begin{document}

\title{Self-Attention Based Semantic Decomposition in Vector Symbolic Architectures}
\author[1]{Calvin Yeung}
\author[1]{Prathyush Poduval}
\author[1]{Mohsen Imani}
\affil[1]{Department of Computer Science, University of California, Irvine}
\affil[ ]{\{chyeung2, ppoduval, m.imani\}@uci.edu}
\date{}

\maketitle

\begin{abstract}
Vector Symbolic Architectures (VSAs) have emerged as a novel framework for enabling interpretable machine learning algorithms equipped with the ability to reason and explain their decision processes. The basic idea is to represent discrete information through high dimensional random vectors. Complex data structures can be built up with operations over vectors such as the ``binding'' operation involving element-wise vector multiplication, which associates data together. The reverse task of decomposing the associated elements is a combinatorially hard task, with an exponentially large search space. The main algorithm for performing this search is the resonator network, inspired by Hopfield network-based memory search operations. 

In this work, we introduce a new variant of the resonator network, based on self-attention based update rules in the iterative search problem. This update rule, based on the Hopfield network with log-sum-exp energy function and norm-bounded states, is shown to substantially improve the performance and rate of convergence. As a result, our algorithm enables a larger capacity for associative memory, enabling applications in many tasks like perception based pattern recognition, scene decomposition, and object reasoning. We substantiate our algorithm with a thorough evaluation and comparisons to baselines.
\end{abstract}

\section{Introduction}
\label{sec:intro}
Modern machine learning (ML) algorithms have succeeded in performing a wide variety of tasks like image recognition, automated driving, recommendation, and drug prediction. Through utilizing large datasets and complex networks, today's ML models have reached very high levels of accuracy. However, one of the key problems faced by machine learning models is their ability to reason. Large Language Models (LLM), for example, can perform well in natural language conversations with their users. However, they perform poorly when faced with tasks involving logical reasoning and simple mathematics, often hallucinating facts and answers~\cite{huang2022towards,huang2023large,zhang2023siren,li2023halueval}. Another example is self-driving cars, where the reasoning process needs to be robust and verifiable so that it does not make any mistakes that may lead to the loss of life. Therefore, equipping ML models with the ability to explain their reasoning in an interpretable manner is a critical issue that is currently at the forefront of research.

In recent years, Vector Symbolic Architectures (VSA), also known as Hyperdimensional Computing (HDC), has garnered significant attention due to its natural similarities with cognitive data structures~\cite{kleyko2023survey}. HDC employs high-dimensional holographic vectors with attributes mirroring those found within brain-based architectures to represent information and perform computation~\cite{kanerva2009hyperdimensional}. Central to the operational efficacy of HDC are three fundamental operations: bundling, binding, and permuting. These mechanisms are pivotal for emulating the essential combinatory patterns of cognitive architectures, namely variable binding, sequential structuring, and hierarchical organization~\cite{frady2020resonator,fodor1988connectionism}. The HDC framework thus facilitates the representation and manipulation of complex data structures in a transparent and interpretable manner, enabling models to maintain and express the intricacies of hierarchical and relational data similarities. This underpins the application of symbolic logical reasoning within cognitive computational models and excels in many perception and logical reasoning tasks.

The primary method of representing information with different attributes in HDC is through associative binding. Here, each attribute is assigned a series of vectors (for each possible value of the attribute), and a particular data point is represented by the element-wise product of the vectors of its attributes. For example, an image containing the number ``7'', with the color blue, and of medium size, is represented by the vector $\vec{H}_{\rm blue, medium, 7}=\vec{N}_7* \vec{C}_{\rm blue}* \vec{S}_{\rm medium}$, where $\vec{N}_i,\vec{C}_j,$ and $\vec{S}_k$ are the vectors representing the number, color and the size attributes respectively. This binding procedure ensures that each unique combination of attributes is represented in a nearly orthogonal manner (as long as the underlying attributes are also represented by nearly orthogonal random vectors). If there are multiple numbers in the same image, the information is memorized and represented through their element-wise sum $\sum_{l}\vec{H}_l$, such that the resulting vector is highly similar to all the attribute combinations it contains, and is dissimilar with all non-existing combinations. Such a framework has been used in solving a wide variety of tasks in a holographic manner, like DNA pattern matching~\cite{poduval2021cognitive,zou2022biohd}, holographic function computations~\cite{poduval2021stochd,poduval2021robust,poduval2022adaptive}, and face detection~\cite{imani2022neural}.

Recent works have leveraged this HDC structure to attain state-of-the-art performance in neuro-symbolic reasoning tasks like the Raven's Progressive Matrices (RPM)~\cite{hersche2023neuro}. The RPM task consists of a series of panels containing polygons of different shapes, numbers, sizes, and colors, and the problem is to predict the pattern followed by the independent attributes. \cite{hersche2023neuro} tackled the task by training a model that can directly output an HDC memorized vector $\vec{H}=\sum_l \vec{H}_l$ containing the information about all the possible shapes in the series of panels. The memory vector was then decomposed into its corresponding attributes. In the RPM tasks, the search space over all the attributes is small, and thus the decomposition over the corresponding attributes can be done through a simple exact search. However, for larger and more general problems, the search space scales exponentially as $\sim N^A$, where $A$ is the number of attributes and $N$ is the number of possible values for each attribute. Such an exact search can also be prohibitive when some attributes (like the size, or opacity) can take a continuous spectrum of values. 

In the current HDC framework, the decomposition of $\vec{H}$ into its set of constituent attributes is performed using the resonator network~\cite{frady2020resonator,kent2020resonator,langeneggerInmemoryFactorizationHolographic2023}. It follows an iterative procedure based on an in-superposition-search framework, where the candidate attributes are first initialized to the average attribute vectors. By then multiplying all the average attribute vectors, the resonator network performs a similarity-based search and projection for each attribute, iteratively, until convergence. This method resembles memory retrieval in a Hopfield network, where the states are stored in an energy functional. Through an iterative energy-minimization algorithm, the Hopfield network is able to retrieve stored states and is supported by rigorous mathematical guarantees on the storage capacity and convergence probability. 

A recent work has shown an interesting equivalence between certain classes of Hopfield networks and self-attention models. By including a norm-bounded energy term to the Hopfield energy functional, it was shown that the iterative update rule for lookup in the memory is equivalent to self-attention applied to the set of memorized vectors \cite{ramsauerHopfieldNetworksAll2021}. Moreover, the resulting memory capacity is exponential in dimension, as opposed to the linear capacity of the traditional Hopfield network. In this work, we make use of the synergy between resonator network and Hopfield network to construct a self-attention based resonator network. Our contributions are as follows:

\begin{itemize}
    \item We construct a new update rule for the resonator network, inspired by the attention-based memory recall rule for the continuous Hopfield network. We show that this update rule enables us to use the resonator network with continuous factors rather than bipolar factors.
    \item We show that the resulting network has high robustness against error, and can store exponentially many combinations of codebook factors. In particular, we demonstrate that the attention-resonator network is highly robust against cross-correlation noise for both bipolar and continuous vectors, while the original resonator network fails with very low accuracy (near zero) for continuous values.
    \item We perform a thorough numerical analysis on the convergence rate, accuracy, and complexity, to demonstrate that our proposed resonator network is scalable and is suitable for practical neurosymbolic tasks, and discuss example use cases.
\end{itemize}

\section{Background}
In this section, we give an overview of Vector Symbolic Architectures (VSAs) and describe how resonator networks work. 

\subsection{Overview of Vector Symbolic Architectures}
Vector Symbolic Architectures (VSAs), also known as Hyperdimensional Computing (HDC), is a computational paradigm based on an algebra over high dimensional vectors. It is a brain-inspired framework as it is motivated by the observation that the brain operates on high dimensional representations \cite{kanervaHyperdimensionalComputingIntroduction2009}.

At the core of HDC are randomly sampled high dimensional vectors, also called hypervectors, representing discrete units. They can be compared via a similarity function $\delta(H_1,H_2)$. Each class of elementary symbols (which can denote different values of an attribute, for example) is represented by a codebook, consisting of a list of high-dimensional, holographic, and random hypervectors that preserve a predefined similarity between different symbols within the class. The two common ways of sampling hypervectors are the bipolar representation and Fourier Holographic Reduced Representation (FHRR). The similarity in both encoding is inherited from the complex Euclidean norm, defined as 
\begin{align}
    \delta(\vec{H},\vec{G}) = \vec{H}^\dagger \vec{G}/D,
\end{align}
where the $\dagger$ denotes the complex-conjugate transpose.

In the bipolar representation, the hypervectors are randomly sampled from $\{-1,1\}^D$, and the hypervectors that represent different values of the same attribute are considered to be mutually orthogonal to each other. The FHRR representation, on the other hand, operates over a $d-$dimensional continuous feature space with feature vector $\vec{f}$, to construct high-dimensional encoding that preserves a kernel similarity $K(\vec{f}-\vec{g})$ over the underlying feature space. This is done using the kernel trick~\cite{rahimi2007random}, where a random $D\times d$ dimensional matrix $W$ is sampled from a probability distribution $p(\omega)$, and then the encoding is performed as $\vec{H}_{{f}}= \exp( i W\vec{f})$. If $p(\omega)$ is chosen as the Fourier transform of a translation invariant kernel $K(\vec{f}-\vec{g})$, then the similarity of hypervectors approximates the underlying kernel as 
\begin{align}
    \delta(\vec{H}_{f},\vec{H}_g)\approx K(\vec{f}-\vec{g}).
\end{align}
Thus, in cases where the attributes can take continuous values (like size, position, or shading), the FHRR representation can be used for encoding data. The set of all possible vectors for a given attribute is called the $\textit{codebook}$. Given hypervectors $x_1,...,x_n\in\mathbb{H}^D$ representing values of some attribute, where $\mathbb{H}^D$ denotes the space in which the hypervectors live, the corresponding codebook is a matrix $\mathbf{X}\in\mathbb{H}^{n\times D}$ such that the $j$-th row of $\mathbf{X}$ is $x_j$.

The three main operations in HDC, bundling, binding, and permutation, can be characterized by how they affect the similarity of hypervectors. We describe the three operations below:
\begin{enumerate}
    \item Bundling ($+$): Typically implemented as element-wise addition. If $H=H_1+H_2$, then both $H_1$ and $H_2$ are similar to $\mathcal{H}$. From a cognitive perspective, it can be interpreted as memorization~\cite{kanerva2009hyperdimensional}.
    \item Binding ($*$): Typically implemented as element-wise multiplication. If $H=H_1*H_2$, then $H$ is dissimilar to both $H_1$ and $H_2$. Binding also has the important property of similarity preservation in the sense that for some hypervector $H_3$, $\delta(H_3*H_1,H_3*H_2)\simeq\delta(H_1,H_2)$. From a cognitive perspective, it can be interpreted as the association of concepts~\cite{frady2018theory}.
    \item Permutation ($\rho$): Typically implemented as a rotation of vector elements. Generally, $\delta(\rho(H),H)\simeq 0$. Permutation is usually used to encode order in sequences.
\end{enumerate}

Using the simple formalisms above, HDC can be used to construct many different types of data structures like trees~\cite{frady2020resonator}, graphs~\cite{poduval2022graphd}, and finite state automata~\cite{osipov2017associative}. One of the most common frameworks in the application of HDC is the ``bind-and-bundle'' scheme~\cite{poduval2023quantum,poduval2023quantum2}, where attributes of each object are first bound together, and then the resulting vector of multiple objects are bundled to represent the object combination. The final vector is similar to only those vectors that represent the constituent objects, which can be created only through the specific binding of attributes.  For example, if a scene contains a hexagon and pentagon with attributes such as size, and shading, then the encoding would be 
\begin{equation}
\begin{aligned}
    h_{\text{scene}} & = h_{hexagon} * h_{large} * \cdots * h_{light} \\
    & + h_{pentagon} * h_{large} * \cdots * h_{dark}.
\end{aligned}
\end{equation}

As a result, HDC is a candidate for addressing the binding problem of neural networks~\cite{rosenblatt1962principles} to analyze real-world scenarios represented by hierarchical and nested compositional structures~\cite{rachkovskij2001binding, hersche2023neuro}. In one such application, \cite{hersche2023neuro} proposes a neural-vector-symbolic architecture (NVSA) consisting of (1) a neuro-vector frontend to generate HDC-like representation from an input scene consisting of different shapes, and (2) a vector-symbolic backend to perform symbolic reasoning and pattern recognition. Moreover, NVSA, like many other HDC models~\cite{poduval2022graphd, nunes2022graphhd,paxon2021computing, gayler2009distributed, osipov2022hyperseed, kim2018efficient, imani2018hierarchical, imani2019quanthd, thomas2021theoretical}, uses the bind-and-bundle framework. 

Many HDC use cases required the decomposition of memorized hypervectors into their elementary components along with the corresponding attributes. This process required some knowledge of the codebook and the encoding process. However, in cases where the codebooks are known, then the decomposition process (through an exact search) has an exponentially large search space size. The resonator network was introduced~\cite{frady2020resonator,kent2020resonator} as a heuristic search process, inspired by the memory recall procedure of the bipolar Hopfield network. It is the current t\textit{state-of-the-art}, and recurrent network that showed practical advantages in decomposing a bound hypervector over optimization-based approaches. However, as we show below, this method is highly susceptible to noise, and the accuracy reduces drastically if the vectors are made continuous instead of bipolar.

\subsection{Traditional Resonator Networks}
For simplicity, we consider the two-codebook case, but the description can be easily extended to an arbitrary collection of codebooks. Given two codebooks $X=\{x_1,...,x_n\}$ and $Y=\{y_1,...,y_m\}$ and some composite vector $s=x_k*y_l$ for some unknown $1\leq k\leq n$ and $1\leq l\leq m$, the goal of a resonator network is to identify $k$ and $l$, i.e. to factorize $s$.

A traditional resonator network operates only on bipolar vectors; i.e. $x_i,y_j\in\{-1,1\}^D$ for all $1\leq i\leq n$ and $1\leq j\leq m$. Let $\mathbf{X},\mathbf{Y}$ be matrices whose rows consist of the elements of the codebooks $X,Y$, respectively. We start by initializing $\hat{x}_0=\frac{1}{n}\sum_{i=1}^n x_i$ and $\hat{y}_0=\frac{1}{m}\sum_{i=1}^m y_i$. We then iterate
\begin{align}\label{eq:res-update1}
\begin{split}
    \hat{x}_{t+1}&=\mathrm{sgn}(\mathbf{X}\mathbf{X}^\top (s*\hat{y}_t)) \\
    \hat{y}_{t+1}&=\mathrm{sgn}(\mathbf{Y}\mathbf{Y}^\top (s*\hat{x}_t))
\end{split}
\end{align}
After a few iterations, $\hat{x}_t,\hat{y}_t$ might converge on the correct values.

The operating principle of the above dynamical system becomes obvious once one understands it as a coupled Hopfield network. Intuitively, $s*\hat{y}_t$ and $\hat{x}_t*s$ give us estimated values for $x_k$ and $y_l$, respectively, given the current best estimates, which are subsequently cleaned up by passing it through the corresponding Hopfield networks.

More generally, for resonator networks with $F$ factors denoted $\hat{x}^{(j)}$ for $j=1,...,F$, we can write the update rule as 
\begin{align}\label{eq:res-update1-general}
    \hat{x}_{t+1}^{(j)}&=\mathrm{sgn}(\mathbf{X}\mathbf{X}^\top (s*\hat{o}_t^{(j)})) \text{ for }j=1,...,F,
\end{align}
where $\hat{o}_t^{(j)}= \Pi_{i\ne j}\hat{x}_t^{(i)}=\hat{x}_t^{(1)}*\cdots*\hat{x}_t^{(j-1)}*\hat{x}_t^{(j+1)}*\cdots *\hat{x}_t^{(F)}$. The intuition for this can be thought of as trying to estimate $x^(j)$ from the previous iteration's estimate, and then unbinding all other factors from the target composite vector $s$.

\subsection{Hopfield Networks}
A Hopfield network is an auto-associative memory model; i.e. it retrieves memory items based on the content of the memory itself. It can be implemented as a neural network that stores patterns $\{\xi_j\}_{j=1}^n$ such that $\xi_j\in \{-1, 1\}^d$ are attractors \cite{hopfieldNeuralNetworksPhysical1982}. An input query $\xi^0$ is passed into the network. The retrieved vector is determined by the update rule
\begin{align}\label{eq:update1}
    \xi^0_{t+1}=\mathrm{sgn}\left(\mathbf{X}\mathbf{X}^\top \xi^0_t\right)
\end{align}
where $X=[\xi_1,...,\xi_n]$. Under conditions such as sufficient separability of patterns (with respect to dot product) and $n$ being sufficiently small, $\xi^0$ will converge to the closest pattern $\xi_j$.

The capacity of the Hopfield network defined above is $O(d)$. \cite{ramsauerHopfieldNetworksAll2021} introduces a new update rule
\begin{align}\label{eq:update2}
    \xi^0_{t+1}=\mathbf{X}\mathrm{softmax}(\beta \mathbf{X}^\top \xi^0_t)
\end{align}
Here, the patterns $\{\xi_j\}_{j=1}^n$ satisfy $\xi_j\in\mathbb{R}^d$. $\beta$ is the inverse temperature hyperparameter. The resulting Hopfield network has exponential storage capacity and has been shown to be equivalent to the self-attention mechanism with some minor adjustments.

\section{Attention-based Resonator Networks}
An attention-based resonator network uses a variant of the update rule in Eq.~\ref{eq:update2} as opposed to that in Eq.~\ref{eq:update1}. Rewriting Eq.~\ref{eq:res-update1-general}, we get
\begin{align}\label{eq:res-update2}
    \hat{x}_{t+1}^{(j)}&=\mathbf{X}\mathrm{softmax}(\beta\mathbf{X}^\top (s*\hat{o}_t^{(j)})/D)\text{ for }j=1,...,F.
\end{align}
Here, $\hat{o}_t^{(j)}=(\hat{x}_t^{(1)})^{-1}*\cdots*(\hat{x}_t^{(j-1)})^{-1}*(\hat{x}_t^{(j+1)})^{-1}*\cdots *(\hat{x}_t^{(F)})^{-1}$. Unlike the original update rule, which only applies to bipolar hypervectors, our proposed attention-based update rule applies to a more general class of hypervectors, including FHRR. We include inverses since by expanding the domain of vectors beyond bipolar hypervectors, hypervectors are in general not self-inverses with respect to element-wise multiplication. Depending on the domain of the hypervectors, we may replace $(\cdot)^\top (\cdot)$ with a more appropriate measure of similarity.

For example, if we were to use codebooks consisting of FHRR hypervectors, the update rule would instead be
\begin{align}\label{eq:res-update3}
    \hat{x}_{t+1}&=\mathbf{X}\mathrm{softmax}(\beta\mathfrak{R}[\mathbf{X}^\dagger (s*\hat{o}_t^{(j)})]/D)\text{ for }j=1,...,F.
\end{align}

\section{Results}
We examine the performance of both traditional and attention-based resonator networks while varying a number of parameters, including codebook size, number of factors, and vector dimension. We characterize performance in terms of accuracy and number of iterations required to reach convergence.

\begin{figure}
    \centering
    \includegraphics[width=0.75\textwidth]{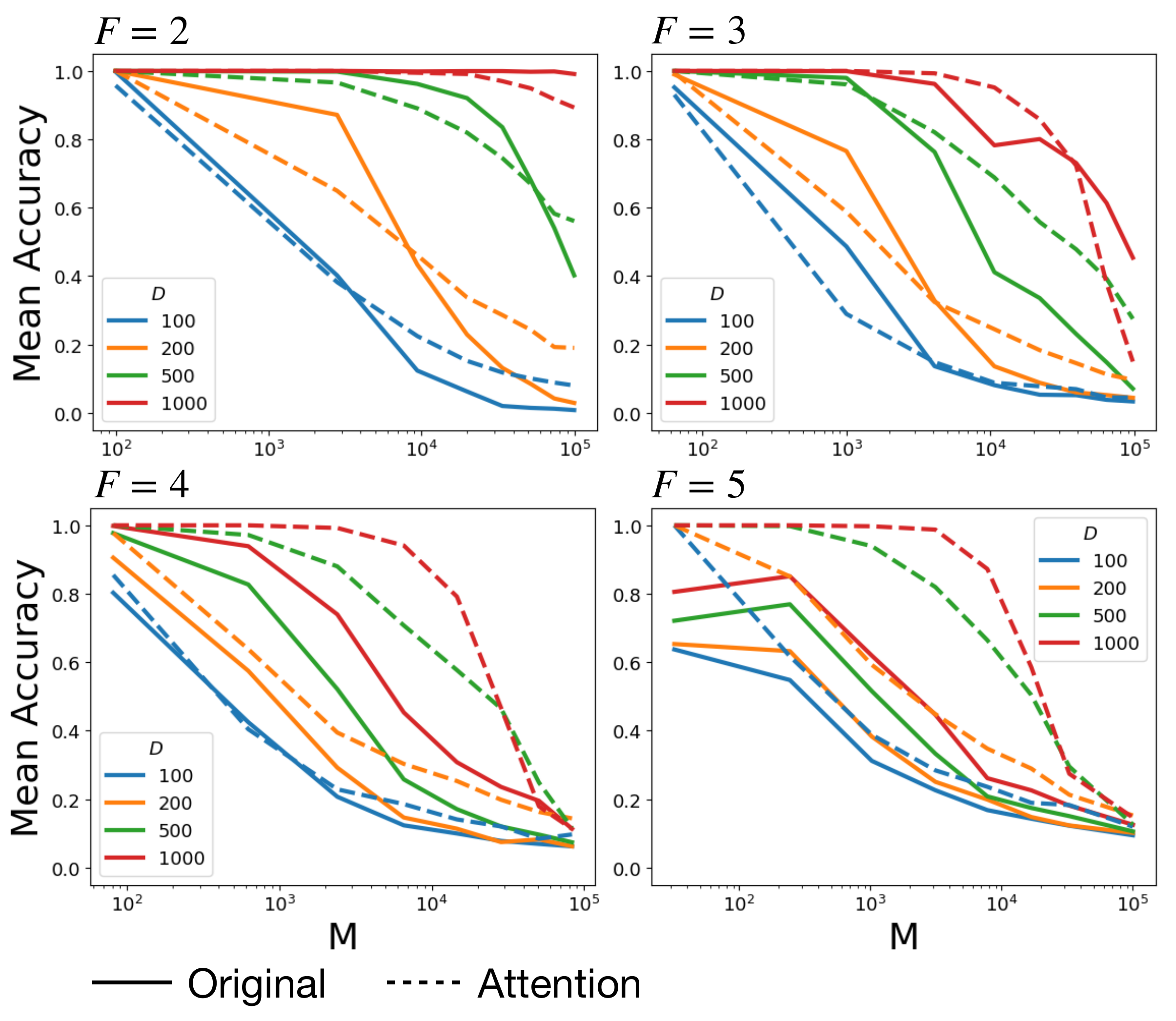}
    \caption{Comparison between resonator networks using the old update rule (solid line) and the new attention-based update rule (dotted line). We vary the size of the search space $M$ and plot the mean accuracy over 1000 trials for different numbers of factors $F$ and vector dimensions $D$. We use bipolar vectors for both resonator networks and $\beta=250$ for the attention-based resonator network.}
    \label{fig:acc-vs-m}
\end{figure}

We randomly generate codebooks $\mathbf{X}_1,...,\mathbf{X}_F$ of size $n$. We choose some arbitrary bound hypervector $s=x_1^{(i_1)}*\cdots *x_F^{(i_F)}$, where $x_j^{(i_j)}$ is the $i_j$-th vector in the $j$-th codebook. \textbf{We iterate the update rule until convergence or until the number of iterations has reached some maximum value $0.001M$ where $M=n^F$ is the size of the search space.} Thus, we investigate the regime in which the number of iterations is much smaller than the total size of the search space so that our results are suggestive of model performance in practical applications. As in \cite{kent2019resonator}, we compute the accuracy as $c/F$ where $c$ is the number of correct factors.

\subsection{Comparison between update rules for bipolar codebooks}
We compare the mean accuracy of resonator networks using the old update rule and the new attention-based update rule as we modulate the search space size $M$ over 1000 trials, as shown in Figure \ref{fig:acc-vs-m}. We use codebooks consisting of bipolar vectors for both resonator networks. We observe while both networks have comparable for $F=2$ factors, the attention-based resonator network outperforms the traditional resonator network for $F>2$, with an increasing gap in performance as $F$ increases.

\begin{figure}
    \centering
    \includegraphics[width=0.75\textwidth]{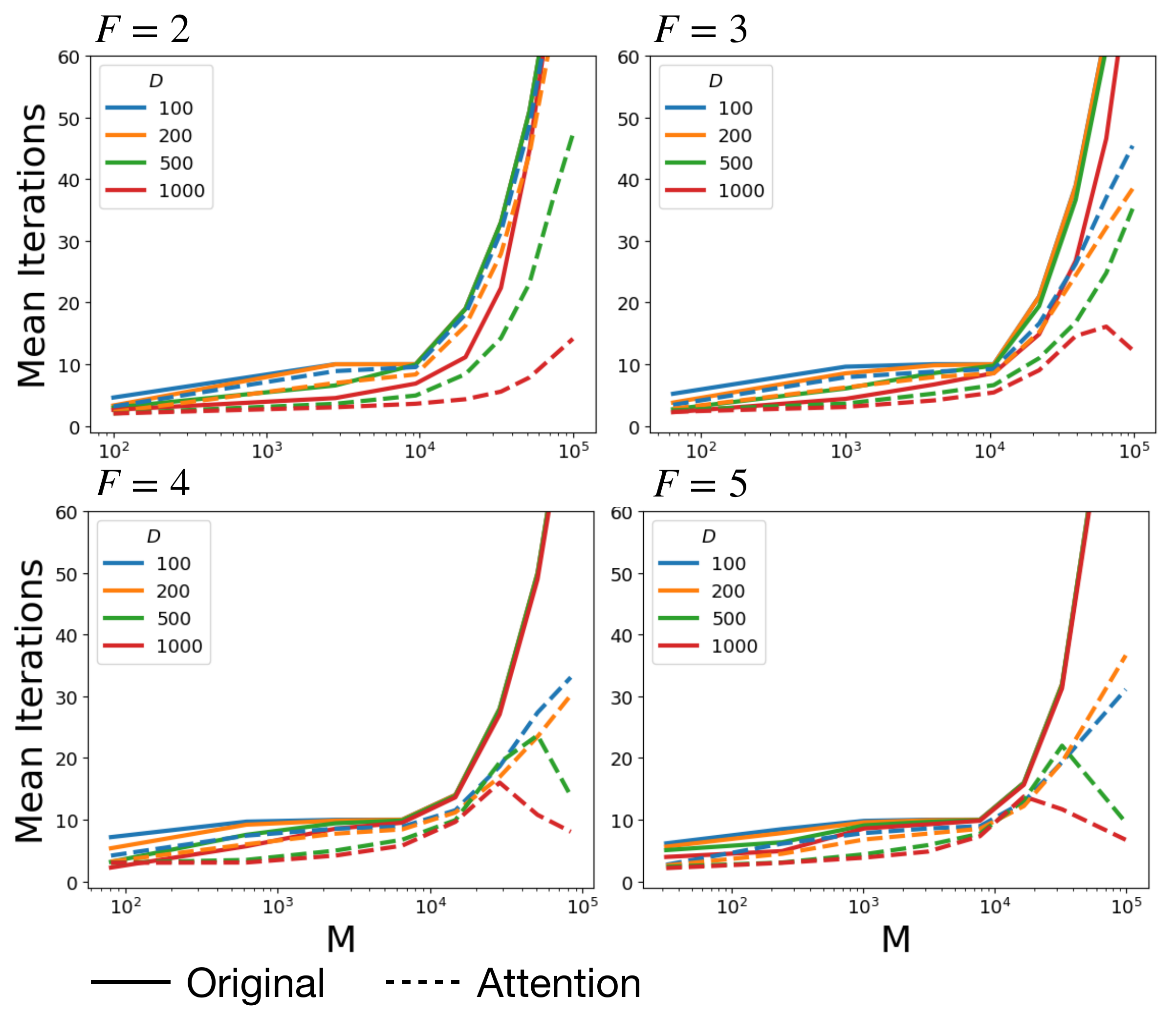}
    \caption{Comparison between resonator networks using the old update rule (solid line) and the new attention-based update rule (dotted line). We vary the size of the search space $M$ and plot the mean number of iterations before convergence over 1000 trials for different numbers of factors $F$ and vector dimensions $D$. We use bipolar vectors for both resonator networks and set the maximum number of iterations to $0.001M$.}
    \label{fig:its-vs-m}
\end{figure}

Figure \ref{fig:its-vs-m} compares the mean number of iterations of resonator networks using the old update rule and the new attention-based update rule as we modulate the search space size $M$. We notice that the attention-based update rule requires significantly fewer iterations to converge, with faster convergence as $F$ increases in general for fixed $M$. This phenomenon occurs as given fixed $M$, larger $F$ implies a smaller codebook size $n$.

On the other hand, the traditional resonator network fails to converge before reaching the maximum number of iterations in general, as their dynamics are known to enter limit cycles \cite{kent2019resonator}.

\subsection{Comparison between update rules for FHRR codebooks}
While the original resonator network only works with bipolar hypervectors, one may adapt it to work for a larger class of hypervectors. Following the update rule in Eq.~\ref{eq:res-update3}, we adapt the update rule in Eq.~\ref{eq:res-update1}  to work for FHRR hypervectors as follows:
\begin{align}\label{eq:res-update4}
    \hat{x}_{t+1}&=g(\mathbf{X}\mathfrak{R}[\mathbf{X}^\dagger (s*\hat{o}_t^{(j)})])\text{ for }j=1,...,F,
\end{align}
where $g$ normalizes each component of the resulting vector $\mathbf{v}\in\mathbb{C}^D$ such that $\|\mathbf{v}_j\|=1$, analogous to $\mathrm{sgn}$, to prevent runaway dynamics. $\hat{o}^{(j)}$ is defined as in Eq.~\ref{eq:res-update2}.

\begin{figure}
    \centering
    \includegraphics[width=0.8\textwidth]{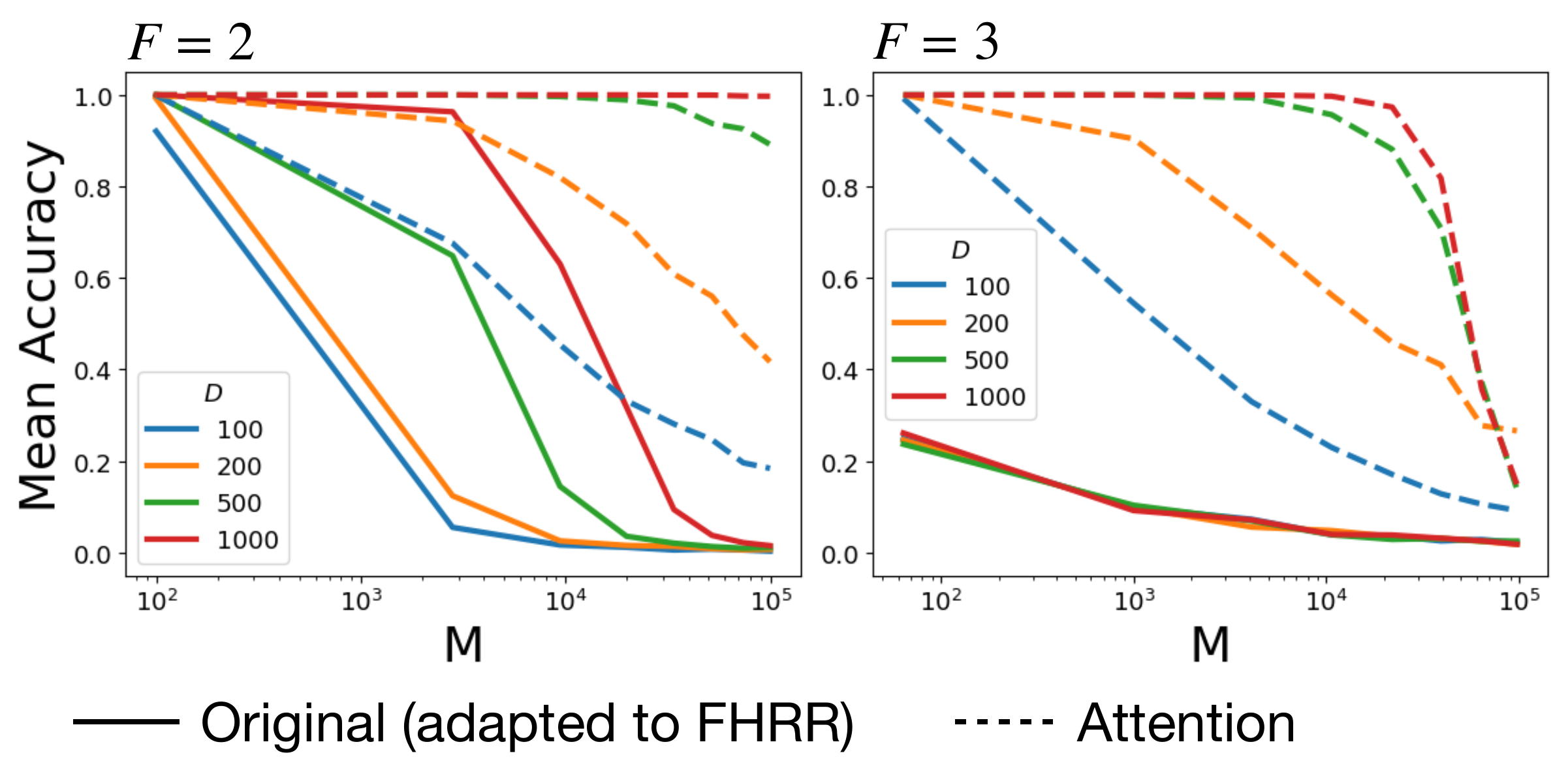}
    \caption{Comparison of mean accuracy between the original resonator network adapted to FHRR (solid line) and the attention-based resonator network (dotted line).}
    \label{fig:acc-vs-m-fhrr}
\end{figure}

Figure \ref{fig:acc-vs-m-fhrr} compares the mean accuracy between the original resonator network adapted for FHRR and the attention-based resonator network for $F=2,3$. The FHRR hypervectors are of the form $e^{i\mathbf{m}}$ with $\mathbf{m}_j\sim\mathrm{Unif}(0, 2\pi)$ for $j=1,...,D$. We observe that the adapted version of the original resonator network significantly underperforms compared to the attention-based resonator network.

\begin{figure}
    \centering
    \includegraphics[width=0.8\textwidth]{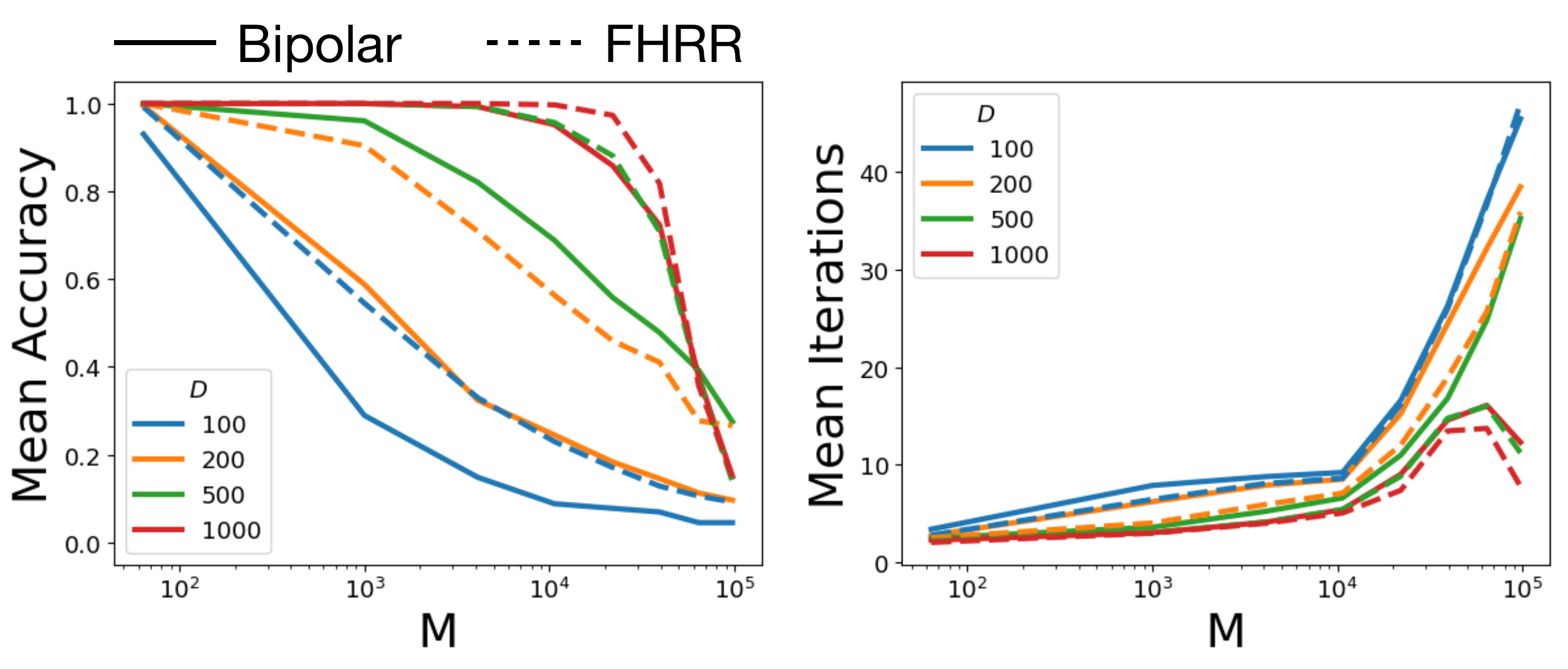}
    \caption{Plots of mean accuracy and mean iterations for attention-based resonator networks using bipolar codebooks (solid line) and FHRR codebooks (dotted line) for $F=3$, over 1000 trials.}
    \label{fig:fhrr-vs-bipolar}
\end{figure}

\begin{figure}
    \centering
    \includegraphics[width=0.7\textwidth]{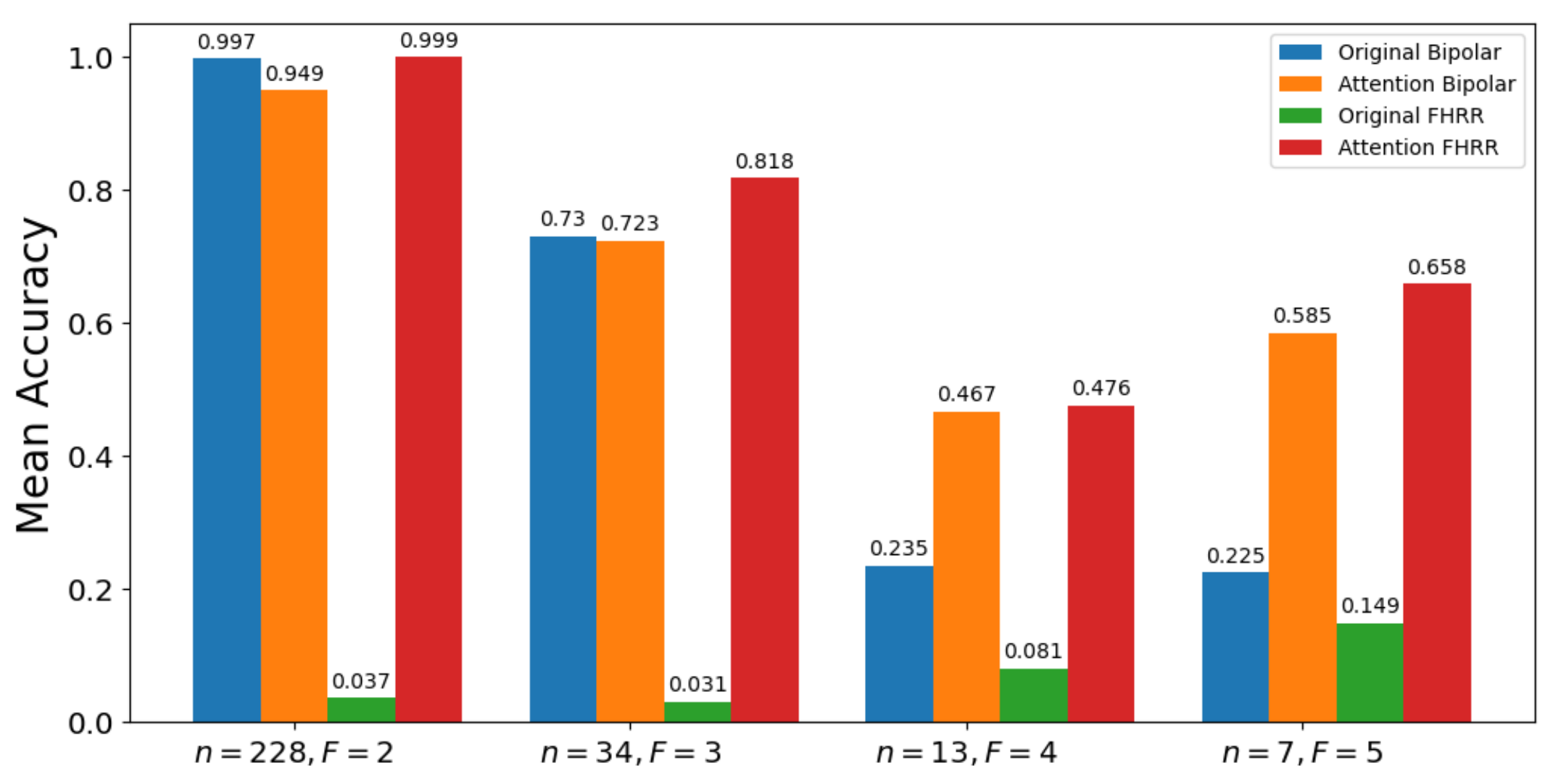}
    \caption{Mean accuracy comparison between original and attention-based resonator networks using both bipolar and FHRR hypervectors over multiple configurations of $n$ and $F$, chosen such that $M=n^F\approx 5000$. Averages are computed over 1000 trials. Attention-based resonator networks use $\beta=250$.}
    \label{fig:bar-comparison}
\end{figure}

\subsection{Comparison between bipolar and FHRR codebooks}
In addition, we may compare the difference between using bipolar and FHRR codebooks while keeping the type of resonator network (i.e. attention-based resonator network) fixed. Figure \ref{fig:fhrr-vs-bipolar} compares the two, visualizing the mean accuracy and mean iterations until convergence as $M$ and $D$ vary. Here, we perform 1000 trials and use $F=3$ and the sample FHRR sampling scheme as in the previous section. The results clearly demonstrate that FHRR provides superior accuracy, with performance comparable to that of the bipolar attention-based resonator network with almost double dimension, while having faster convergence rates at the same time.

Figure \ref{fig:bar-comparison} compares the difference in mean accuracy for the original and attention-based resonator networks using both bipolar and FHRR hypervectors over multiple configurations of $n$ and $F$, chosen such that $M=n^F\approx 5000$. We observe that the attention-based resonator network consistently outperforms all other models, with the difference between attention-based and original resonator networks becoming larger as $F$ increases. 

\begin{figure}
    \centering
    \includegraphics[width=0.65\textwidth]{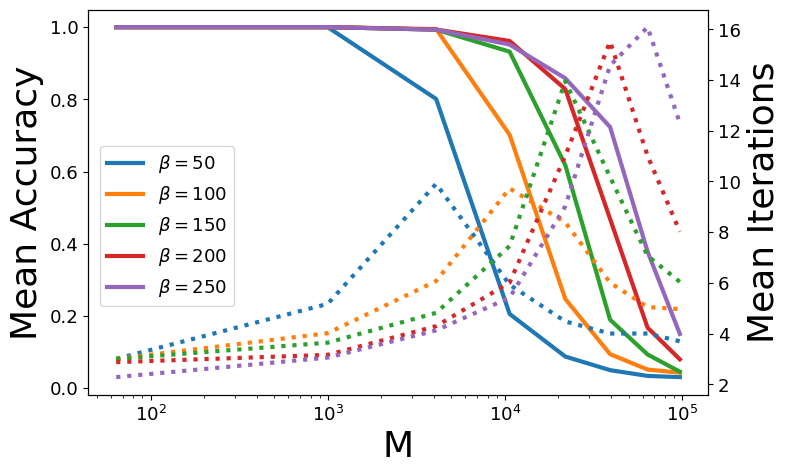}
    \caption{Plots of mean accuracy and mean iterations for attention-based resonator networks for different values of $\beta$, over 1000 trials. Here, $F=3$ and we use bipolar hypervectors with $D=1000$.}
    \label{fig:beta-acc}
\end{figure}

\subsection{Effect of $\beta$ on the attention-based resonator network}
The inverse temperature $\beta$ controls how much weight is placed on a single element in the codebook when the update rule forms a linear combination of codebook vectors. This can have a substantial effect on the dynamics. Figure \ref{fig:beta-acc} visualizes how $\beta$ affects both mean accuracy and mean iterations until convergence. In general, we see consistently improved accuracy with increasing $\beta$ as well as mean iterations.

\subsection{Decoding a bundle of bound hypervectors and noise tolerance}

\begin{figure}
\centering
    \includegraphics[width=1\textwidth]{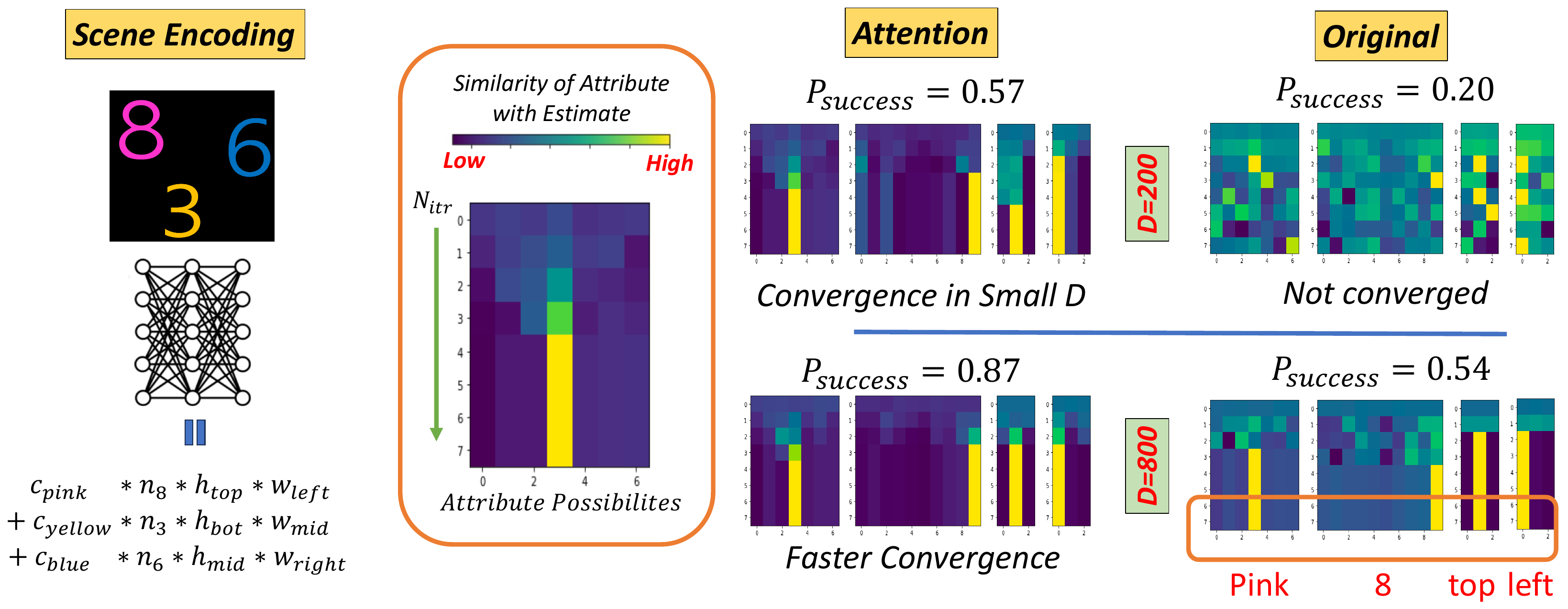}
    \caption{Example use case, as performed in~\cite{poduval2023quantum,poduval2023quantum2}. \textbf{Left:} The scene representation is the bundling of bounded vectors that represent each number and is typically generated either through HDC mathematics or through a neural network. \textbf{Middle:} The output of the resonator network can be visualized as a heatmap for each attribute, where the $x$-axis represents different values of the attributes and the $y$-axis denotes the iteration, and the value of the heatmap denotes the similarity with each possible attribute vector. \textbf{Right:} We show the output of the attention and original resonator network in decoding at $D=200$ and $D=800$, demonstrating the accurate and fast decoding capabilities with the attention update rule. }
    \label{fig:comparison}
\end{figure}

Most applications of HDC involve a superposition of different bound vectors representing different objects in the memory. Suppose we have a bundle of bound hypervectors of the form
\begin{align}
    \begin{split}
        s&=\sum_{j=1}^k x_1^{(i_{1j})}*\cdots *x_F^{(i_{Fj})}
    \end{split}
\end{align}
We can decode $s$ into its constituents by iteratively applying the resonator network. By decoding one constituent element at a time, we can subtract it from $s$ (also referred to as ``reasoning out''~\cite{frady2020resonator}) and apply the resonator network again to decode the remaining parts. An example use case is shown in Figure \ref{fig:comparison}, where we have a scene consisting of numbers of various colors in different locations. The scene representation is the bundling of bound vectors that represent each number. We also show the success rate $P_{success}$, which is the probability with which the network converges onto a correct factorization contained in the bundle of terms. 

\begin{figure}
    \centering
    \includegraphics[width=0.85\textwidth]{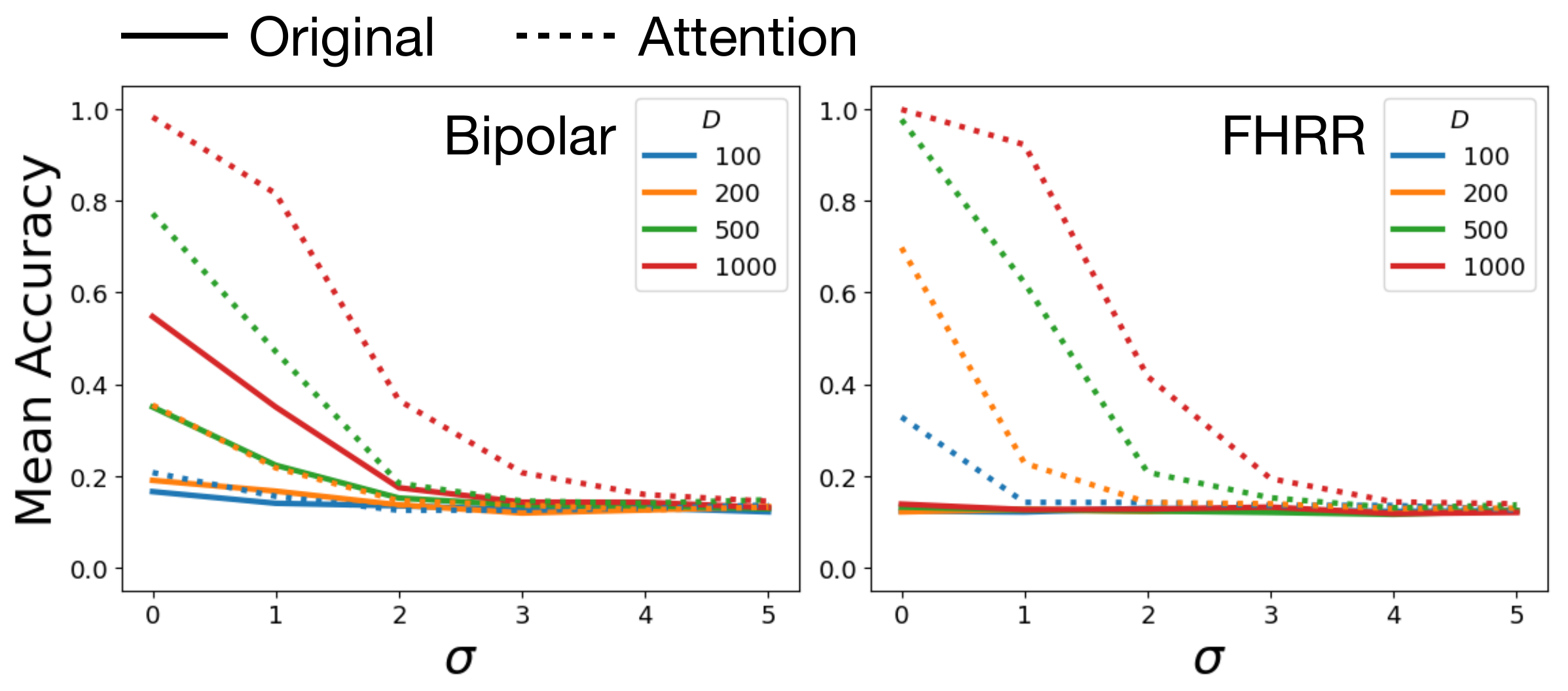}
    \caption{Plot of mean accuracy of the original and attention-based resonator network when noise $\epsilon\sim N(0,\sigma)$ is added to the bound hypervector $s$ for both bipolar (left) FHRR (right) hypervectors using both the original (solid line) and attention-based (dotted line) update rules. Here, $F=4$ and $n=8$. Averages are computed over 1000 trials.}
    \label{fig:acc-vs-noise}
\end{figure}

In low dimensions, we see that the original resonator network does not converge often, with a failure rate of about $1-P_{success}\sim 0.8$. The Attention resonator network, on the other hand, succeeds almost $2/3^{rds}$ of the time. As we increase the dimension to $D=800$, the original succeeds only with probability $0.54$, while the attention method succeeds with $0.87$ probability. 

\begin{figure}
    \centering
    \includegraphics[width=0.75\textwidth]{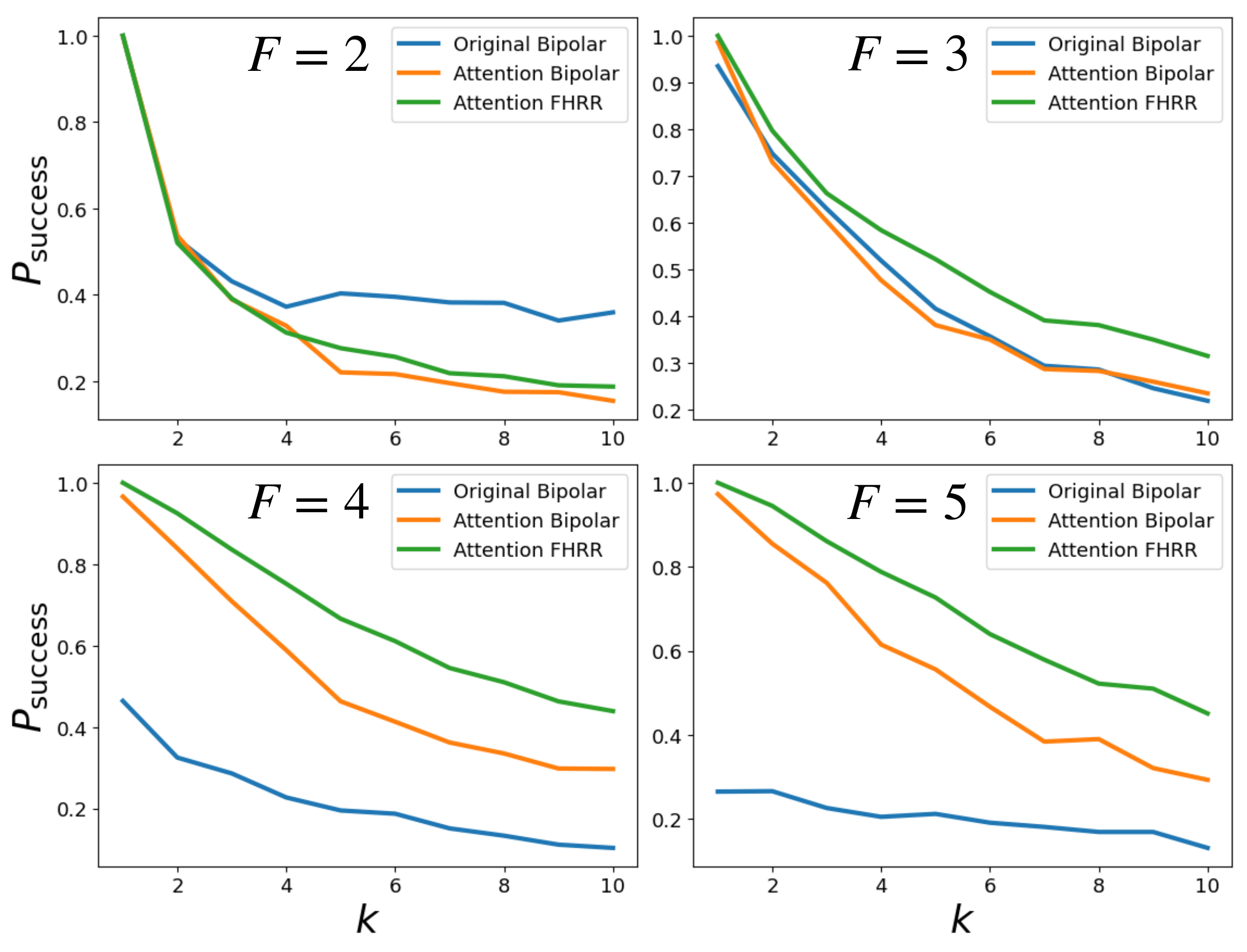}
    \caption{Plots of $P_{success}$ of variants of the resonator network for different numbers of constituents in the bundle $k$ and for different values of $F$. For each plot, $n$ is chosen such that $M=n^F\approx 5000$. Probabilities are estimated over 1000 trials. We note that the FHRR representation with the original network leads to almost zero success, and so we omit its results in the future figures. }
    \label{fig:success-vs-k}
\end{figure}

To thoroughly characterize the effect of the bundled terms, we perform two different analyses. First, we focus on only one single factorization and add a random Gaussian noise with mean $\mu=0$ and standard deviation $\sigma$. (In a bundle of $k$ terms where we focus on a single factorization, in the worst case scenario, the remaining $k-1$ terms can be approximated as Gaussian noise with standard deviation $\sigma\propto\sqrt{k-1}$.) In Figure~\ref{fig:acc-vs-noise}, we show the effect of mean accuracy as we continuously increase the noise. In both the bipolar and FHRR representations, the attention version of the resonator network outperforms the original version by a large margin. 

Next, we calculate the success rate $P_{success}$ of converging on \textit{at least} one correct factorization when we have $k$ different factorizations bundled together. This is shown in Figure~\ref{fig:success-vs-k}, where the success rate drops substantially as a function of $k$. However, in cases of multiple factors, the attention version substantially outperforms the original resonator network.

We summarize our results in Table~\ref{tab:table1} and \ref{tab:table2}, for various configurations of $F$ and $k$. Table~\ref{tab:table1} shows the accuracy, demonstrating that the attention-based FHRR and attention-based bipolar resonator networks outperform the original network by a large margin as the number of factors and number of bundled terms increase. Moreover, the original FHRR-adapted resonator network has an almost $0$ accuracy for all the parameters shown in the table. In Table~\ref{tab:table2}, we show the method complexity, which is defined as the average number of steps required for correct convergence, i.e.
\begin{align}
    {\rm Complexity } = \frac{\langle N_{\rm itr}\rangle}{P_{success}}.
\end{align}
We can see the attention-based FHRR and attention-based bipolar resonator networks have an order of magnitude lower complexity as compared to the original bipolar resonator network. The original FHRR-adapted resonator network has almost $\infty$ complexity since it has close to zero probability of success.

\begin{table}[]
    \centering
    \begin{tabular}{lrrrrrrrrr}
\toprule
            Method (Prob.) &     $(2,1)$ &  $(2,3)$ &  $(2,9)$ &  $(4,1)$ &    $(4,3)$ &     $(4,9)$ &  $(6,1)$ &    $(6,3)$ &  $(6,9)$ \\
\midrule
Attention Bipolar & $0.998$ & $0.377$ & $0.178$ & $0.977$ & $0.708$ & $0.291$ & $0.928$ & $0.780$ & $0.364$ \\
   Attention FHRR & $\mathbf{1.000}$ & $0.367$ & $0.195$ & $\mathbf{0.999}$ & $\mathbf{0.853}$ & $\mathbf{0.465}$ & $\mathbf{0.991}$ & $\mathbf{0.926}$ & $\mathbf{0.584}$ \\
 Original Bipolar & $\mathbf{1.000}$ & $\mathbf{0.433}$ & $\mathbf{0.380}$ & $0.438$ & $0.272$ & $0.106$ & $0.141$ & $0.125$ & $0.099$ \\
    Original FHRR & $0.001$ & $0.000$ & $0.000$ & $0.000$ & $0.002$ & $0.001$ & $0.000$ & $0.002$ & $0.000$ \\
\bottomrule
\end{tabular}
    \caption{Probability of success for original and attention-based resonator networks using both bipolar and FHRR hypervectors over different configurations of $(F,k)$, with $F=2,4,6$ and $k=1,3,9$. The highest probability for each configuration is in bold.}
    \label{tab:table1}
\end{table}

\begin{table}[]
    \centering
    \begin{tabular}{lrrrrrrrrr}
\toprule
            Method (Compl.) &     $(2,1)$ &  $(2,3)$ &  $(2,9)$ &  $(4,1)$ &    $(4,3)$ &     $(4,9)$ &  $(6,1)$ &    $(6,3)$ &  $(6,9)$ \\
\midrule
Attention Bipolar &     $3.10$ &  $\mathbf{9.03}$ & $19.06$ &  $4.75$ &    $6.41$ &    $15.67$ &  $6.11$ &    $6.56$ & $13.00$ \\
   Attention FHRR &     $\mathbf{3.01}$ &  $9.22$ & $\mathbf{18.78}$ &  $\mathbf{4.49}$ &    $\mathbf{5.32}$ &     $\mathbf{9.91}$ &  $\mathbf{5.58}$ &    $\mathbf{5.70}$ &  $\mathbf{8.39}$ \\
 Original Bipolar &     $5.37$ & $23.03$ & $26.32$ & $19.02$ &   $32.67$ &    $85.35$ & $62.87$ &   $71.42$ & $87.03$ \\
    Original FHRR & $\infty$ &   $\infty$ &   $\infty$ &   $\infty$ & $\infty$ &  $\infty$ &    $\infty$ & $\infty$ &   $\infty$ \\
\bottomrule
\end{tabular}
    \caption{Complexity for original and attention-based resonator networks using both bipolar and FHRR hypervectors over different configurations of $(F,k)$, with $F=2,4,6$ and $k=1,3,9$. The lowest complexity for each configuration is in bold.}
    \label{tab:table2}
\end{table}

\section{Conclusion}
We propose a new attention-based update rule for the resonator network and show that this update rule enables us to use the resonator network with continuous factors rather than bipolar factors, resulting in a network that has superior performance in terms of accuracy and speed of convergence, in the regime where the number of iterations is much smaller than the total size of the search space, in particular for factorization problems with a larger number of factors. In addition, we show that the resulting network has high robustness against error which leads to superior performance in decomposition problems with bundle size greater than one. We perform a thorough numerical analysis on the convergence rate, accuracy, and complexity, and our results suggest that our proposed model is scalable and is suitable for neurosymbolic tasks that require symbolic decomposition.

\section{Acknowledgements}
This work was supported in part by the DARPA Young Faculty Award, the National Science Foundation (NSF) under Grants \#2127780, \#2319198, \#2321840, \#2312517, and \#2235472, the Semiconductor Research Corporation (SRC), the Office of Naval Research through the Young Investigator Program Award, and Grants \#N00014-21-1-2225 and \#N00014-22-1-2067. Additionally, support was provided by the Air Force Office of Scientific Research under Award \#FA9550-22-1-0253, along with generous gifts from Xilinx and Cisco.

\bibliographystyle{unsrt}
\bibliography{main}
\end{document}